\documentclass[letterpaper]{article}
\usepackage{aaai}
\usepackage{times}
\usepackage{helvet}
\usepackage{courier}

\usepackage{graphicx}
\usepackage{float}
\usepackage{dcolumn}
\usepackage{longtable}
\usepackage{booktabs} 
\usepackage{hyperref}
\usepackage[utf8]{inputenc}
\usepackage[english]{babel}
\usepackage{amsthm}

\theoremstyle{plain}
\newtheorem{thm}{Theorem}

\theoremstyle{definition}
\newtheorem{defn}[thm]{Definition} 
 
\usepackage{hyperref}
\hypersetup{
    urlcolor=cyan,
}
 
\begin{document}
%
\title{About Face: A Survey of Facial Recognition Evaluation}
\author{

Inioluwa Deborah Raji\\
\And
Genevieve Fried\\
}

\maketitle

\begin{abstract}
\begin{quote}
We survey over 100 face datasets constructed between 1976 to 2019 of 145 million images of over 17 million subjects from a range of sources, demographics and conditions. Our historical survey reveals that these datasets are contextually informed - shaped by changes in political motivations, technological capability and current norms. We discuss how such influences mask specific practices - some of which may actually be harmful or otherwise problematic - and make a case for the explicit communication of such details in order to establish a more grounded understanding of the technology's function in the real world. 
\end{quote}
\end{abstract}

\section{Introduction}

Whether in schools \cite{shultz_2019}, convenient stores \cite{spears_2019}, public squares \cite{bridges_2019,mesnik_2018,coolfire_2019}, concerts \cite{bridges_2019}, apartment complexes \cite{durkin_2019}, airports \cite{oflaherty_2019}, neighbourhood parks \cite{chinoy_2019}, or on personal devices \cite{apple_support_2019},  facial processing technology (FPT) is increasingly pervading our lives in numerous, unaccountable ways. Earlier this year, the National Institute of Standards (NIST) proudly announced that between 2014 and 2018, FPT improved twenty fold, to a failure rate of just 0.2 percent \cite{nist_2018}. 

Yet a string of failed real world pilots contradicts the academic mythos of facial recognition as a solved problem. Eight trials of FPT deployments in London between 2016 and 2018 resulted in a 96\% rate of false identifications as criminal suspects \cite{dearden_2019}. A 2019 report found that 81\% of suspects flagged through the facial recognition tool used by London’s Metropolitan Police were wrongly identified \cite{manthorpe_martin_2019}. Similarly, New York City’s Metropolitan Transportation Authority (MTA) abandoned a facial recognition pilot after the technology failed to properly identify anyone (it had an 100\% error rate) \cite{berger_2019}. Furthermore, these failures are not evenly distributed across demographic subgroups. A study in 2018 revealed that for gender classification, commercial facial recognition API’s performed up to 30\% worse on a darker skinned female subgroup compared to  a lighter male subgroup \cite{Joy1}. A follow up audit in 2019 \cite{AA}, as well as subsequent studies by NIST \cite{ngan_grother} and other academics \cite{vangara2019characterizing}, have confirmed these disparities and demonstrated their extension to other problems such as face identification and verification tasks. Similarly, a year after Amazon Rekognition systems were shown to falsely match 28 congress members \cite{ref48},  the technology falsely matches 27 mostly minority athletes to criminal mugshots \cite{Refsports}, drawing particular public attention to  the limitations of this technology in deployment. 

Aside from functional concerns, several cities have responded to the threat presented by the use of FPT for surveillance by banning its use completely by government actors \cite{noauthor_community_nodate,administrative_code,haskins_oakland_2019,noauthor_surveillance_2018,noauthor_ordinance_nodate} and others have sought to restrict the use of the technology in certain deployment contexts \cite{montgomery_precision_nodate}, such as housing \cite{clarke_text_2019} or in schools \cite{nys_biometrics_bill}. Many states have passed laws specifically addressing the privacy violations inherent in the development and operation of FPT systems \cite{washington_5528,idaho-0118,texas_business_commerce,ab1281-privacy,ab1215-facial,arkansas-1943,newyork-assembly,illinois_sb1719,illinois-740}, with many more states presiding over legislative proposals \cite{washingon_5376,michigan0342,ab1215-facial,masachusetts_1385,arizona2478,florida-1270}, and a federal bill pending \cite{blunt_s.847_2019}. 
On the federal side, the Commercial Facial Recognition Privacy Act of 2019  prohibits ``certain entities from using facial recognition technology to identify or track an end user without obtaining their affirmative consent purposes'' \cite{blunt_s.847_2019}. 


Yet despite the growing public awareness of the practical failure of these systems once released in the real world, academic studies continue to report near perfect performance of facial recognition systems on benchmark datasets. 
In an attempt to better understand this dissonance between the perceived functionality of these systems under current narrow evaluation norms and the reality of its overall holistic performance when deployed, we surveyed over 100 datasets from the recorded beginnings of digital facial processing technology in the 1960s to present day.
We analyze the evolution of evaluation tasks, data and metrics to gain a clearer picture of what will be required for evaluations to truly capture a reliable representation of the performance of these systems in a deployed context. 


This is the largest and most recent survey of this kind that the authors are aware of - the last survey of this kind was conducted in 2012 \cite{forczmanski2012comparative} with only forty-one datasets; prior to that, a survey was conducted in 2005 with just twenty-one datasets \cite{jain2011handbook}. 


\section{Terminology \& Scope}
Facial processing  technology (FPT) in this study will be considered as a broad term to encompass any task involving the identification and characterization of the face image of a human subject. This includes face detection——the task of locating a face within a bounding box in an image, face verification——the one-to-one confirmation of a query image to a given image, face identification——the one-to-many matching of a query image to the most similar results within a given repository of images, and facial analysis——a classification task to determine facial characteristics, including physical or demographic traits like age, gender or pose, as well as more situational traits such as facial expression. 

Mainstream commercial facial recognition products are still predominantly based on still 2-D image-based predictions \cite{wang2018deep} so we limit the scope of this survey to the consideration of 2-D still-image photographic facial recognition benchmarks that are presently available online. This omits datasets comprised of non-visual face images representing infra-red or other sensor output maps, sketch or drawing datasets, video-based datasets, 3-D image datasets and datasets addressing full body human identification. 



In this study, we refer to an audit or an evaluation as any process that is used to determine the suitability of a particular technology to fulfil its intended use in the specified context of deployment. The method is thus independent of this definition and thus an evaluation may include qualitative and qualitative approaches. In particular, a “black box” audit refers to an external evaluation done by an independent third-party on a system which remains inaccessible or unknown to the auditor. 

\section{Historical Context of Facial Recognition Development}

We begin with a brief overview of the historical context of FPT development in order to anchor our understanding of the major shifts that defined the evolution and technical progress of this technology, which in turn shapes the norms of evaluation for this technology. 

We divide our historical survey into four periods defined by three key turning points of facial recognition development: (1) the creation of the Face Recognition Technology (FERET) database in 1996, the very first large scale face dataset available for academic and commercial research \cite{Ref27}, (2) the Labeled Faces in the Wild (LFW) dataset in 2007, as the first Web-sourced and unconstrained face dataset \cite{LFWTech}, and (3) the development of DeepFace in 2014, the first facial recognition model to beat human performance on the face verification task and to be trained with the now-dominant technique of deep learning \cite{taigman2014deepface}. 

\subsection{Period I: Early Research Findings (1964 - 1995)}
In 1964, Woodrow Bledsoe  first attempted “facial recognition” in a computational form. Funded by an “undisclosed intelligence agency” and armed with a book of mugshots and a probe photograph, he used a computer program to connect the identity of the suspect to an identity in the book of mugshots \cite{bledsoe1966model}. Success in this criminal identification task was measured by the ratio of the number of guesses required to identify the true match to the query image over the total number of faces in the dataset.

Bledsoe's initial approach was to encode each individual with a vector of computed distances between facial features, a method that would become popular but was very computationally expensive and slow -  with the technology at the time, Bledsoe could only process around 40 pictures an hour \cite{bledsoe1966model}. Eventually, a new method called eigenfaces, which represented the pixel intensity of face features in a lower dimensional space, offered an appealing alternative approach. Yet obtaining enough data at the time to attempt such new methods was challenging, as researchers had to recruit and hire models and photographers, manually design the set up for consistent or controlled illumination, and manually label data, including facial landmarks \cite{jafri2009survey}. 

\subsection{Period II: Commercial Viability as the “New Biometric” (1996 - 2006)}

By 1996, government officials had recognized and embraced the face as a non-invasive biometric attribute that could be used to track and identify individuals without requiring their explicit physical participation \cite{phillips2000introduction}.
The Face Recognition Technology (FERET) dataset was thus created with \$6.5 million of funding from the U.S. Department of Defense and the National Institute of Standards and Technology (NIST) to provide researchers the data they required to make progress in the field. In 15 photography sessions of the same set up between August 1993 and July 1996, images were collected in a semi-controlled environment \cite{Ref27}. The resulting benchmark began with 2,413 still face images, representing 856 individuals, and grew to contain 14,126 facial images of 1,199 individuals, available upon request. At the moment of its release, it became the largest and most comprehensive effort to create a benchmark that would accurately compare and evaluate existing facial recognition algorithms \cite{Ref27}. The large data effort coupled with a government sponsored effort to promote facial recognition algorithm development via competitions and research investments \cite{FRGC} proved successful at igniting academic research interest in the field. 

In 2000, given the success of the FERET database in stimulating research interest in facial recognition, commercial implementations of this technology began to appear and prompted the National Institute of Standards and Technology (NIST) to release the Facial Recognition Vendor Test (FVRT), a benchmark aimed at evaluating this emerging commercial systems. Even then, the expressed intended context of consideration for these tools were to be “applied to a wide range of civil, law enforcement and homeland security applications including verification of visa images, de-duplication of passports, recognition across photojournalism images, and identification of child exploitation victims” \cite{FRVT2015}.

The creation of a larger,  more substantial dataset allowed early computer vision methods, such as support vector machines (SVMs), simple convolutional neural networks (CNNs) and hidden Markov models (HMMs), to be  applied to facial recognition with some promising results \cite{jafri2009survey}. However, the commercialization attempts with these early methods revealed that even small environmental changes, such as in image illumination and a subject’s pose, could at this time be enough to obscure or distort the features required to make a match. Similarly, any unexpected change in their face - from aging to a new facial expression to  partial occlusions, such as a scarf, mask, or pair of glasses – could cripple the performance of the technology \cite{sharif2017face,forczmanski2012comparative,yang2002detecting}.   



Given the dearth of available face data, certain strategies to better generalize across environments were still out of reach and it was considered that, at this time, “current algorithms for automatic feature extraction do not provide a high degree of accuracy and require considerable computational capacity” \cite{jafri2009survey}.

\subsection{Period III: Mainstream Development for Unconstrained Settings (2007-2013)}

The development of the Labeled Faces in the Wild (LfW) dataset addressed researchers' desire to have access to a more naturally situated and varied data. The dataset leveraged the Web to source the first fully unconstrained face dataset with over 13,000 images of 1,680 faces in an infinite combination of poses, illumination conditions, and expressions \cite{LFWTech}. 

LFW inspired a flurry of Web-scraped face datasets for facial recognition model training and benchmarking - including many datasets sourcing images without consent from online platforms, such as Google Image search \cite{bainbridge2013intrinsic,han2017heterogeneous,cao2018vggface2}, Youtube \cite{chen2017spoofing,dantcheva2012can},  Flickr \cite{2019arXiv190110436M,kemelmacher2016megaface} and Yahoo News \cite{jain2010fddb}. As the appetite for unstructured and unconstrained “in the wild” data grew, there was also in this period a proliferation of benchmarks like ChokePoint \cite{wong2011patch} and SCface \cite{grgic2011scface},  datasets that source face images from mock or real surveillance set ups. 
As datasets began to more closely resemble real world conditions, so did the evaluations of commercial products. The Facial Recognition Vendor Test (FVRT) evolved extensively over this period \cite{frvt2000,frvt2002,FRVT2015}, growing from 13,872 images of about 1,462 subjects in the initial implementation in 2000 to 30.2 million still photographs of 14.4 million individuals in the iteration in 2013. 

The research problem of identifying faces in unconstrained conditions nevertheless  remained a stubborn technical challenge and development stalled as academics struggled to develop methods to represent faces independently of a controlled image context and template appearance. 

\subsection{Period IV: Deep Learning Breakthrough (2014 and onwards)}

It was not until the breakthrough of Alexnet in 2012, and the subsequent introduction of the DeepFace model in 2014, that the use of neural networks became a mainstream method for facial recognition development. DeepFace, the first facial recognition model trained with deep learning, was also the first instance of a facial recognition model approaching human performance on a task. Deepface was developed by researchers at Facebook, Inc. and trained on an internal dataset composed of images from Facebook profile images; at the time, it was purportedly “the largest facial dataset to-date, an identity labeled dataset of four million facial images belonging to more than 4,000 identities”  \cite{taigman2014deepface}.The impact of deep learning techniques on face recognition and its adjacent problems was dramatic; the DeepFace model achieved a  97.35\% accuracy on the Labeled Faces in the Wild (LfW) test set, reducing the previous state of the art’s error by 27\%.

In response to this technological advance, the size of subsequently constructed face datasets grew significantly to accommodate the growing data requirements to train deep learning models. The rapid progress sparked high commercial interest, as well. Moving beyond security applications, facial recognition products began to encompass use cases that include “indexing and searching digital image repositories”, “customized ad precise delivery”, “user engagement monitoring” and “customer demographic analysis”\cite{FRGC}. In the search for datasets sufficient to use in the training and testing of these data-hungry methods, many were inspired by Web-sourced benchmarks such as LfW, resulting in datasets such as Oxford’s VGG-Face dataset \cite{vgg}, Microsoft’s 1M MS Celeb \cite{guo2016ms}, MegaFace \cite{kemelmacher2016megaface},  and the CASIA WebFace dataset \cite{yi2014learningcasia}.  

After the 2014 breakthrough of the DeepFace model’s human-level performance on facial recognition, there was a shift made to commercialize the technology. In 2015, NIST launched the IARPA Janus Benchmark-A face challenge (IJB-A) , which was an open challenge in which researchers executed algorithms on NIST-provided image sets, and returned output data to NIST for scoring. The competition was organized by the Intelligence Advanced Research Projects Activity (IARPA), an organization within the Office of the Director of National Intelligence. This challenge and its variations and updates IJB-B and IJB-C ran from 2015 to the end of 2017, growing into a dataset of 138,000 face images, 11,000 face videos, and 10,000 non-face images of celebrities and Internet personalities collected from the web. The  3,531 subjects included in the dataset are specifically designed not to overlap with the popular face recognition benchmarks at the time, such as VGG-Face and CASIA WebFace dataset, in order to prevent overfitting.

\section{Survey of Facial Recognition Evaluation}

We execute a historical survey of 133 datasets created between 1976 to 2019. The datasets collectively representing 145,143,610 images of 17,733,157 individual people’s faces. Celeb 500k of 2018 is the largest dataset, containing 50,000,000 images \cite{cao2018celeb}, and the FRVT Ongoing challenge data from NIST contains the most image subjects, including the faces of 14,400,000 \cite{frvtongoing}. The smallest dataset is 54 images of 4 people from 1988’s JACFEE (Japanese and Caucasian Facial Expressions of Emotion) dataset \cite{biehl1997matsumoto}. Overall, on average here are 1,262,118 images and 159,758 subjects. 

We then do a chronological analysis of these currently accessible face datasets. We note trends in the design decisions made with the release of these benchmarks and datasets  and map how such trends feed into or result in current misunderstandings of the limitations of this technology upon deployment. The full details of the datasets included in the survey can be found linked \href{https://docs.google.com/spreadsheets/d/1ZA_Ijr0bFfXHVzlpxA1JkimD-V6ZdGDJkim4fWgenNQ/edit?usp=sharing}{\underline{here}}\footnote{Additional breakdown of the dataset details can be found here: https://tinyurl.com/shbraqn}. A quantitative summary of results for each period can be seen in Table 1. The following is a summary of high-level findings.

\begin{table*}
  \small
  \centering
  \label{data_summary}
  \caption[Historical Arcs of Facial Recognition Development]{Historical Arcs of Facial Recognition Development.}
  \begin{tabular}{lrrrr}
    \toprule
Period & Period I  & Period  II & Period III  & Period IV \\
    \midrule
    Years & Before 1996 &  1996 - 2007 &  2007-2014 &  After 2014 \\
    \midrule
    Number of Datasets Created & 5 & 37 & 33 & 45 \\
\midrule
    Range of number of Images in a dataset (MIN- MAX) & 56 - 14,126 & 120 - 121,589 & 154 - 750,000 & 642 - 50,000,000 \\
    
    \midrule
    Range of number of Subjects in a dataset (MIN- MAX) & 4 - 1,199 & 10 - 37,437 & 32 - 40,395 & 50 - 14,400,000 \\
    
    \midrule
    Average number of images in a dataset & 2,032 & 11,250 & 46,308 & 2,620,489 \\
    
        \midrule
    Average number of subjects in a dataset & 136 & 1,641 & 4,078 & 75,726 \\
    
    \bottomrule
  \end{tabular}
\end{table*}

\subsection{Task Selection}


Tasks are highly influenced by who is creating and funding the dataset. At times, especially for government datasets, the goal of the developed technology is explicit and specifically defined in the design of the evaluation - for instance, the NIST FRVT dataset is funded by the Department of Homeland Security and contains data sourced from “U.S. Department of State’s Mexican non-immigrant Visa archive”\cite{frvt2002}. The prioritized and dominant use case for this technology is thus still security, access control, suspect identification, and video surveillance in the context of law enforcement and security \cite{sharif2017face,zhao2003face}. We can see from the historical context that the government promoted and supported this technology from the start for the purpose of enabling criminal investigation and surveillance. 

More diverse applications, such as the integration into mobile devices, robots, and smart home facility User Interfaces \cite{wang2018deep}, monitoring user engagement or social objectives such as finding missing children \cite{chexia_2015} emerge only in Period IV. Facial analysis tasks emerge only in the most recent period as well. The exceptions to this are emotion datasets, which, with much older benchmarks,  are often datasets sourced from the Psychology field and re-purposed as evaluations of machine learning models.

Over time, these models were no longer released as complete software packages, but are now deployed as Application Program Interfaces (APIs), providing a pre-trained model-as-a-service that can be integrated into any developer application. This means that facial recognition models are now accessible to any developer seeking to apply the model to their particular use case. Models are thus now widely deployed and embedded into a variety of software products used in unknown and unpredictable contexts. 

Outside of these reported and realized use cases, companies selling facial recognition speak of alternate applications in the marketing copy released with their products. The language in this online marketing copy seems to pivot from being targeted at a broad range of businesses, appealing to uses for advertising and content moderation. These advertised use cases include “indexing and searching their digital imagine barriers”, “customized ad precise delivery”, “user engagement monitoring” and “customer demographic analysis” \cite{FRGC}. There are also mentions of government-compatible interests, such as “face-based user verification” and “security monitoring”, though the expected context of use is left ambiguous and is worded as though aimed to be used as corporate security tools or part of commercial products rather than in law enforcement or for government purposes.

Facial analysis is the class of tasks that is likely to include the most ambiguous model objectives, often implicating the “discredited pseudosciences of physiognomy and phrenology” \cite{metz_facial_2019}, where a subject’s inner state is wrongly inferred through the evaluation of that subject’s external features. Pseudo-scientific tasks to predict “ sexual orientation” \cite{wang_kosinski,leuner_replication_2019}, “attractiveness” \cite{eisenthal_facial_2006,schmid_computation_2008}, “hireability” \cite{fetscherin_effects_2019}, “criminality” \cite{wu_automated_2016}, and even more accepted but contested attributes, such as affect \cite{picard_affective_2000}, gender \cite {keyes_misgendering_2018} and race \cite{benthall_racial_2019}, are rarely questioned in the evaluation of the system. The potential for certain tasks or use cases to cause harm are not often considered or reflected on explicitly during system testing. 


Following the introduction of Amazon's Mechanical Turk service in 2005, researchers began making heavy use of the service in an attempt to clean and make sense of their data, while also enabling the datasets to be used to address additional tasks \cite{irani2013turkopticon}. Certain data and meta-labels sourced for the images are controversial. For instance, the CelebA dataset contains five landmark locations, and forty binary attributes annotations per image. These labels include the problematic and potentially insulting labels regarding size -  “chubby”, “double chin” - or inappropriate racial characteristics such as “Pale skin”
“Pointy nose”, “Narrow eyes” for Asian subjects and “Big nose” and “Big lips” for many Black subjects. Additionally there is the bizarre inclusion of concepts, such as “bags under eyes”, “5 o’clock shadow” and objectively impossible labels to consistently define, such as “attractive” \cite{CelebA}.


\subsection{Benchmark Data}
 Face data benchmarking practice has historically been shaped by the needs of stakeholders most influential in driving model development. Although face data is biometric information as unique and identifiable as a fingerprint, it is also casually available in many forms and can thus be passively collected in ways likely to perpetuate severe privacy violations. 

\subsubsection{\textit{Dataset Size}}


After the release of DeepFace in 2014 \cite{taigman2014deepface}, the demonstration of the effectiveness of deep learning prompted a growing belief in the need for larger scale datasets in order to satisfy the data requirements of such methods. Datasets grew from tens of thousands of images to millions in likes of MegaFace and VGG-Face2. The goal was to create datasets large enough dataset to avoid overfitting and have enough of a variance to be meaningful, yet also of acceptably data quality \cite{wang2018deep}. One can also aim to set up a benchmark with depth, which has a limited number of subjects but many images for each subjects (such as VGGFace2 \cite{VGGface2})  or a dataset  breadth, meaning the set contains many subjects but limited images for each subject (such as MS-Celeb-1M \cite{guo2016ms} and Megaface \cite{kemelmacher2016megaface}. 




\subsubsection{Data Sources}


When  data requirements for model development were low, the common practice was to set up \emph{photo shoots} in order to capture face data controlled for pose, illumination, and expression. Subject consent for participation and data distribution as well as photo ownership are often mentioned explicitly in references for datasets with photography data sources. Depending on the scale of these projects, producing high quality datasets in this vein was highly expensive. And for such a set up, details like camera equipment specifications would matter in determining the quality of the image and overall dataset.

As an alternative, datasets were also sometimes a \emph{collection} curated from other image datasets perhaps built for a different purpose, or simply \emph{crowdsourced} from willing participants who donated their face data after being convinced or paid to do so. Many government collection sources for face data include specifically \emph{mugshots}, often of “deceased persons with prior multiple encounters” \cite{founds2011nist}. In addition to this, stills from webcam footage and official documentation such as VISA photos \cite{FRVT2015}. 


Later academic and corporate sources tended to derive more from the Web \cite{kemelmacher2016megaface,vgg,LFWTech,guo2016ms} 
through \emph{web searches} for still-image examples of “unconstrained” faces, or by taking frames from online videos. Some databases also tapped \emph{surveillance camera} footage to mine face data \cite{grgic2011scface,duke_ristani2016performance,brainwash_stewart2016end}.  
In these cases, the cameras were a set up in a cafe, school campus or public square \cite{duke_ristani2016performance,brainwash_stewart2016end}
- effectively a more subversive photo shoot to capture "in the wild" data. Either case can often be seen as a violation of subject consent.

The diversification of data sources from photography sessions to more crowdsourced and Web-based data sources allowed for a greater diversity of subjects and image conditions, all at a much lower cost than previous attempts. However, in exchange for more realistic and diverse datasets, there was also a loss of control, as it became unmanageable to obtain subject consent, record demographic distributions, maintain dataset quality and standardize attributes such as image resolution across Internet-sourced datasets. 



\begin{figure}
\centering
\includegraphics[scale=0.35]{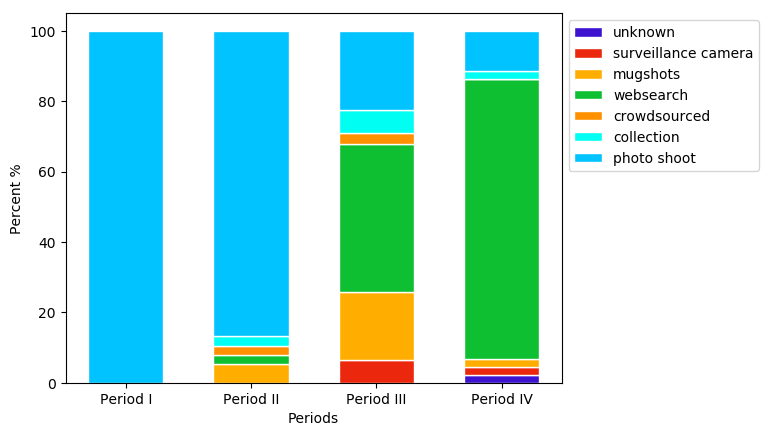}
\caption{Data source by distribution.}
\centering
\label{data_count}
\end{figure}

\begin{figure}
\centering
\includegraphics[scale=0.35]{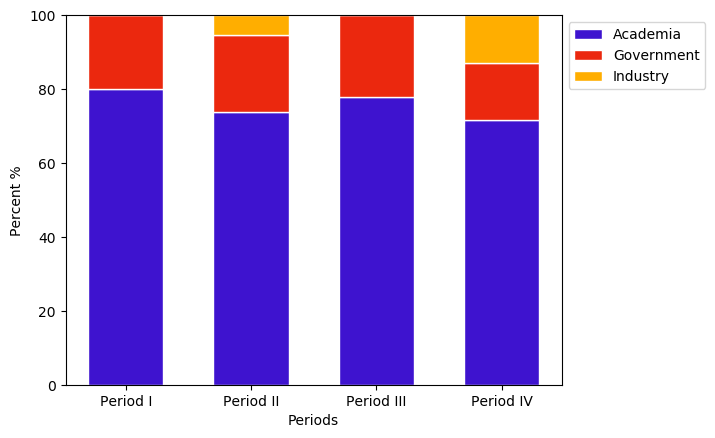}
\caption{Creator types by distribution.}
\centering
\label{creator_type}
\end{figure}

\subsubsection{\textit{Data Sharing}}


Datasets constructed from photo shoots in Period I and II pay considerable attention to issues of copyright and the protection of image ownership rights in distribution practices, with papers for benchmark datasets from these periods often indicating the informed consent of individuals participating in a photoshoot, and including a custom privacy policy \cite{Ref27}. For example, the report describing the FRVT 2000 challenge benchmark dataset comments: “The subjects appearing in the images are all unpaid volunteers who had been briefed on the purpose of their participation and who had positively consented to the study. For privacy reasons the data was gathered anonymously” \cite{frvt2000}. However, the distribution of the datasets from this period was often physically restricted anyways, as several academic datasets required sending images via physical hard drives, incurring a cost to distribution that disappeared with the shift to online options \cite{biehl1997matsumoto,Ref27}.

Once datasets became accessible online, consent and privacy became difficult to manage. Certain strategies, such as sharing hyperlinks rather than the downloaded images, de-anonymizing identities, restricting online dataset access or aiming to crawl the photos of only celebrities and public figures emerged in later eras to address this challenge. The level of awareness of privacy concerns seemed to differ greatly - at times, unconsenting adult subjects were included in datasets available for direct online download \cite{bainbridge201210k}, and other times, distribution of this biometric information was handled sensitively, completely closed off to the public and evaluated exclusively through a custom API or graphic user interface, such as NIST's Biometric Experimentation Environment (BEE) test environment. However, many situations are somewhere in between both extremes, with dataset access granted following a formal request and agreement to the presented terms of use. Data disclosure and distribution practice can also be culturally specific. For instance, the Iranian dataset \cite{iranian_facedatabase} includes female subjects but specifies not to allow for the public display and distribution of this female subgroup specifically, likely due to cultural restrictions of their exposure.

\subsubsection{\textit{Dataset Reporting}}

The reporting of datasets is wildly unstandardized. Many datasets lack information about the source and methodology by which images are collected, or fail to include information at the macro (e.g. demographic) and micro (e.g. image specific attributes or metadata creating) level, producing an incomplete picture of the dataset characteristics. Datasets might be described in an academic paper and/or on a project website, with no standardized format of disclosure, and potential inconsistencies even across different communication mediums and references. For instance, in several cases, the number of images reported on a website might differ from the number in the published paper - and at times both numbers could contradict the size reported in a survey paper or subsequent study working with the dataset \cite{forczmanski2012comparative}. This indicates a lack of provenance and reporting norms to track and appropriately communicate about face dataset construction and evolution.

Interestingly, some of the most comprehensive reporting was performed by NIST as part of their series of FRVT challenges, which are ongoing. Evaluation reports meticulously document the construction (source and method of collections) of their benchmark data. They acknowledge the importance of doing so in their 2000 evaluation report: “Image collection and archival are two of the most important aspects of any evaluation. Unfortunately, they do not normally receive enough attention during the planning stages of an evaluation and are rarely mentioned in evaluation reports.” \cite{frvt2000} 

\subsubsection{\textit{Demographic Representation}}


Although a recently revived topic of interest, researchers flag the propensity for racial bias in the FRVT dataset \cite{frvt2000}, and even indicate model performance disparities over gender and age as early as 2002 \cite{frvt2002}. Understanding the existence of the issue, a surprising number of datasets, especially in Period I and Period II report the limitations of the demographic distributions of the presented dataset, with some even choosing to focus the entire dataset on one demographic, such as Asian-Celeb \cite{asian_celeb_dataset}, Iranian Faces \cite{iranian_facedatabase} and Indian Faces\cite{indian_face_database2018}. Online sourced  datasets seemed to shift towards a Western media default for demographic representation, and, as the datasets were so large, the phenomenon was difficult to track and had been until recently largely unreported \cite{2019arXiv190110436M}. 
Several datasets have been built in response to recent awareness of this issue in order to specifically address the dearth of diversity in mainstream facial analysis benchmarks \cite{Wang2018RacialFI,age-bias-mitigate,2019arXiv190110436M,Joy1}.

While datasets typically have accompanying documentation that describes the types of categories in a dataset (e.g. pose, illumination, etc), only a subset of them explicitly state information about demographics present in the dataset, and far fewer explicitly communicate the exact numbers of images for a given demographic. 

Most notable are the Pilot Parliaments Benchmark \cite{buolamwini2018gender}, which splits demographics across fitzpatrick skin types, as well as the DiveFace dataset, which is expressly annotated to “train unbiased and discrimination-aware face recognition algorithms” yet divide human ethnic origins into only three categories \cite{morales2019sensitivenets}.

In some cases, the hunt for increased demographic diversity may result in inappropriate privacy violations. This is most evident with the LfW+ dataset \cite{han2017heterogeneous}, where Google Image search results for keywords such as “baby”, “kid”, and “teenager” were used to identify juvenile images to supplement to mainly adult subjects of Labeled Faces in the Wild (LfW) \cite{LFWTech}. This dataset and others - from NIST's CHEXIA dataset \cite{flanagan2015chexia}, to CAFE, a child affective dataset \cite{lobue2015child} - rarely involve even the parental consent of involved parties, putting juveniles at risk by exposing their sensitive biometric information.



\subsection{Evaluation Criteria}
The way in which the evaluation process of a model occurs embeds certain insights as to what makes a particular approach to evaluation reliable and more widely influential in defining the norms of evaluation practice in facial recognition.   

\subsubsection{\textit{Consistency of Results}}

In order for an audit to be reliable, there needs to be a guaranteed consistency to the benchmarks being used - both in terms of ethical expectations and standards, as well as the data itself. Data consistency can become especially difficult with the introduction of Web-sourced data, as urls become obsolete. Inconsistencies in ethical expectations and performance standards can also make comparison difficult from year-to-year. One element of process that is yet to be standardized is auditing scheduling - currently there is no timing mentioned as a key component of audit procedure, and without the anticipation of a regular audit period then there is no expectation for regular compliance with expectations.  

Updates to equipment such as digital cameras can affect benchmark attributes such as data resolution. Within our survey, the range of photo sizes and resolutions across benchmarks is large - from 32x32 to 3072x2048 or even larger. As the number of pixels constitutes the direct input to methods such as deep learning, it becomes difficult to understand which element of reported performance metrics are dependent on these other variables. 

\subsubsection{\textit{Metrics}}

As facial recognition tasks evolved from verification and identification to facial analysis, the underlying technical problem evolved from an image similarity search task to a classification task. Such a shift of bucketing test examples into categories can become challenging when considering the limits of even our demographic categories for gender \cite{Keyes:2018:MMT:3290265.3274357} and race \cite{Benthall:2019:RCM:3287560.3287575}.  

There are effectively two groups of evaluations - that of a biometric evaluation for facial recognition and face identification tasks as well as that of classification accuracy for facial analysis tasks. The biometric matching process resembles image similarity search and ranking as a task, and metrics are anchored to a binary output of a match or no match. Meanwhile, classification is really about the assignment of a test example to a class category that matches the pre-determined ground truth label. 

For biometric evaluation, the outcome is binary. Given two predictive outcomes - negative (ie. no match) or positive (ie. a match), we designate N to be all negatively predicted outcomes and P to be all positively predicted outcomes. If a negative prediction is true, it becomes a "True negative" (TN) result, otherwise we can designate it a "False negative" (FN) result. Similarly, if a positive result is correct, it becomes a "True Positive" (TP), counter to a "False Positive" (FN) if such is not the case. False Match Rates (FMR), and False Non-Match Rates (FNMR) are the primary metrics used for facial recognition evaluation, and are at times reported across a range of decision thresholds. 

The details for these calculations in their provided mathematical definitions are as follows.

\begin{defn} \textbf{ False Match Rate (FMR) - Type I error.}
Deciding that two biometrics match, when they do not constitutes a false match. The the frequency of this occurrence is the False Match Rate (FMR) or Type I error, defined as follows:
  \begin{displaymath}
   FMR = FP / N = FP / (FP + TN)
  \end{displaymath}
  
\end{defn}

\begin{defn} \textbf{ False Non-Match Rate (FNMR) - Type II error.}
Deciding that two biometrics do not match, when indeed they do constitutes a false non-match. The frequency of this occurrence is the False Non-Match Rate (FNMR) or Type II error, defined as follows: 
  \begin{displaymath}
  FNMR = FN / (FN + TP) 
 \end{displaymath}
  
\end{defn}


\begin{defn} \textbf{Classification Accuracy.}
Given data set $D = (X,Y)$, where $y_i$ is the ground truth label of a given sample input $d_i$ from $D$, we define black box classifier $f : X, Y \mapsto c$, which returns a prediction $c$ from the attributes $x_i$ of a given sample input $d_i$. We thus define classification error to be as follows:
    \begin{displaymath}
    Acc = P(g(x_i,y_i) = C_i)
    \end{displaymath}
Accuracy can also be stated with respect to binary outcomes, given the provided definitions of "True negative" ($TN$), "False negative" ($FN$), "True Positive" ($TP$), and "False Positive" ($FN$).

 \begin{displaymath}
    Acc = (TP+TN)/(TP+TN+FP+FN)
    \end{displaymath}
\end{defn}

The framing of evaluation calculations can actually become quite political, with various institutions at various points re-framing reported calculations in order to appear better performing. A common practice is to modify the confidence threshold required to make a positive prediction in order to manipulate reported metrics, rather than reporting the area under the receiver operating characteristic curve (AUC), or calibrating metrics at a fixed threshold. In fact, when confronted by researchers about the poor performance of their facial recognition product on certain demographics \cite{AA}, Amazon explicitly manipulates accuracy reporting by re-assessing their product on a higher threshold to claim better performance, even though their police clients were operating at the much lower default threshold \cite{gizmodo}. 

\subsubsection{\textit{Community Adoption}}
Another thing to consider is the level of community adoption of a particular data benchmark and its influence on facial recognition development. Collectively, the analyzed face datasets are known to be cited at least 74,211 times - implying an incredibly wide reach. Not every dataset included in our survey had an available accompanying paper, so it was not possible to obtain citations for our entire survey. The most cited dataset in our survey is the FERET dataset, developed by NIST in 1996 \cite{Ref27}. The factors contributing to why a certain dataset becomes the dominant cited benchmark for a particular period remain uncertain. It can be assumed that government sponsored benchmarks, often tied to a competition or opportunity for academic funding, incentives attention from the research community. Pioneering benchmarks and datasets of particularly high quality are also likely to garner more citations. For instance, Labeled Faces in the Wild, as the first benchmark to make use of Web images and include face data in natural environments, has had enduring relevance.


\begin{table}
  \small
  \label{data_cite}
  \caption[Most Influential Face Datasets per Era]{Most Influential Face Datasets per Era.}
  \begin{tabular}{lrrr}
    \toprule
Period & Dataset Name &Citations & Year Created \\
    \midrule
   Period I & Picture of Facial Affect & 5,163 & 1976 \\
    \midrule
   Period II & FERET & 8,126 & 1996 \\
       \midrule
   Period III & Labeled Faces in the Wild & 3,746 & 2007 \\
       \midrule
   Period IV & VGGFace & 2,547 & 2015 \\
    \bottomrule
  \end{tabular}
\end{table}

\subsubsection{\textit{Qualitative Assessments}}
There is an opportunity to include holistic evaluations of the product and fold that into a larger audit process. The FVRT developed by NIST, for instance, was a two-part audit process involving a “Recognition Performance Test”, which was a quantitative assessment of accuracy, and the “Product Usability Test”, involving a more qualitative evaluation of the ease in making use of the system in deployment \cite{FRVT2015}. 

An extension of this concept could be used to record information about the ethical compliance of the audited model’s use. This can encompass consideration for the product’s context of deployment and process of prediction, in addition to reflections on consideration for privacy and cooperation throughout the audit, including the respect for any documentation requirements. 

While the focus of many legislative proposals are centered on the risk of privacy, and the threat of biometric information being passively collected and analyzed without active and informed consent, many facial recognition evaluations do not currently require reporting on the privacy practices involved in the development of an audited tool, or an articulation of even its intended use case.

\section{Recommendations}


Facial recognition evaluation has evolved rapidly over the last few decades, and we are just beginning to understand how these changes impact our understanding of the performance of a facial recognition system upon deployment. 

Over several periods, we've seen the trend in facial recognition evaluation shift broadly from a highly \emph{controlled}, \emph{constrained} and \emph{well-scoped} activity to one that is not. 
As data sources became even more invasive and difficult to manage, the field has been steadily progressing towards the current crisis of ill-advised tasks and datasets we can see today \cite{megapixels}.
Much harm was brought on by the lack of due diligence that occurred as the scale and application scope of these technologies increased significantly. As practitioners sought web-based or surveillance sources to satisfy increasing data requirements for model training, less attention was paid to consent and privacy concerns, and ethical practices such as reporting subject demographics and monitoring data distribution became less practical and thus less common. In the same way, expanding model distribution from purchased software to internet-accessible API calls, took away control from model developers and increased the scope of system influence, at times far beyond the context of the intended use for the model. Furthermore, the current period of dataset creation through image retrieval at scale on the web raises serious concerns around privacy, ownership, and consent--a legal question that computer scientists should also be actively engaging. There is thus a clear need to be more cautious in the development and dissemination of datasets containing such sensitive biometric information, and a required reflection on when it may not be appropriate to source and handle this kind of data at all. 

Additionally, the current level of documentation in evaluation and data management processes seems insufficiently comprehensive. This suggests the need for data reporting standards to be created, particularly given the amount of inconsistency in data reporting.  
Some of the most critical details about a facial recognition system, such as its context of deployment, it's technical limitations and appropriate scope of use, are missing and/or not communicated within the evaluation process in any way.  A more contextual evaluation is  necessary to address and communicate all the risks of this technology and determine if it should be released in society at its current scale of deployment, or at all. 
At minimum, an important intervention moving forward is to standardize documentation practice, of the model and the face datasets meant to be used in development or evaluation. Proposals such as Model Cards \cite{ModCards}, Datasheets for Datasets \cite{gebru2018datasheets}, or the Data Privacy Label \cite{kelley2009nutrition} can provide reporting guidelines to support the development of such practice as the new normal in the field. 

Finally, in this survey, it is made clear that this initial narrow objective for the technology was that of police or military surveillance. From the earliest facial recognition system to modern NIST benchmarks, it is clear that this is a technology that was historically developed for the purpose of identifying suspects for pursuit and apprehension, whether in the context of law enforcement, war or immigration. Despite current attempts to revisit the narrative and re-frame the purpose of the technology to supposedly benign commercial applications, this history has shaped everything from the nature of the data collected for benchmarks to the nature of evaluation metrics, and certainly the definition of tasks. Those working to improve this technology must awknowledge its legacy as a military and carceral technology, and their contribution toward those objectives.




\section{Conclusion}




Facial recognition technologies pose complex ethical and technical challenges. Neglecting to unpack this complexity - to measure it, analyze it and then articulate it to others - is a disservice to those, including ourselves, who are most impacted by its careless deployment. Dataset evaluation is a critical juncture at which we can provide transparency and even accountability over facial recognition systems, and interrogate the ethics of a given dataset towards producing more responsible machine learning development.

\newpage
\begin{quote}
\begin{small}
\bibliographystyle{aaai}
\bibliography{FRT_survey.bib}

\begin{thebibliography}{}

\bibitem[\protect\citeauthoryear{{ACLU}}{2019a}]{noauthor_community_nodate}
{ACLU}.
\newblock 2019a.
\newblock Community {Control} {Over} {Police} {Surveillance}.

\bibitem[\protect\citeauthoryear{ACLU}{2019b}]{Refsports}
ACLU.
\newblock 2019b.
\newblock Facial recognition technology falsely identifies famous athletes.

\bibitem[\protect\citeauthoryear{{Apple Inc.}}{2019}]{apple_support_2019}
{Apple Inc.}
\newblock 2019.
\newblock About face id advanced technology.

\bibitem[\protect\citeauthoryear{Bainbridge, Isola, and
  Oliva}{2013}]{bainbridge2013intrinsic}
Bainbridge, W.~A.; Isola, P.; and Oliva, A.
\newblock 2013.
\newblock The intrinsic memorability of face photographs.
\newblock {\em Journal of Experimental Psychology: General} 142(4):1323.

\bibitem[\protect\citeauthoryear{Bainbridge}{2012}]{bainbridge201210k}
Bainbridge, W.
\newblock 2012.
\newblock 10k us adult faces database.

\bibitem[\protect\citeauthoryear{Bastanfard, Nik, and
  Dehshibi}{2007}]{iranian_facedatabase}
Bastanfard, A.; Nik, M.~A.; and Dehshibi, M.~M.
\newblock 2007.
\newblock Iranian face database with age, pose and expression.
\newblock {\em Machine Vision}  50--55.

\bibitem[\protect\citeauthoryear{Benthall and
  Haynes}{2019a}]{benthall_racial_2019}
Benthall, S., and Haynes, B.~D.
\newblock 2019a.
\newblock Racial {Categories} in {Machine} {Learning}.
\newblock In {\em Proceedings of the {Conference} on {Fairness},
  {Accountability}, and {Transparency}}, {FAT}* '19,  289--298.
\newblock New York, NY, USA: ACM.
\newblock event-place: Atlanta, GA, USA.

\bibitem[\protect\citeauthoryear{Benthall and
  Haynes}{2019b}]{Benthall:2019:RCM:3287560.3287575}
Benthall, S., and Haynes, B.~D.
\newblock 2019b.
\newblock Racial categories in machine learning.
\newblock In {\em Proc. of the Conference on Fairness, Accountability, and
  Transparency (FAT)}.

\bibitem[\protect\citeauthoryear{Berger}{2019}]{berger_2019}
Berger, P.
\newblock 2019.
\newblock Mta's initial foray into facial recognition at high speed is a bust.

\bibitem[\protect\citeauthoryear{Biehl \bgroup et al\mbox.\egroup
  }{1997}]{biehl1997matsumoto}
Biehl, M.; Matsumoto, D.; Ekman, P.; Hearn, V.; Heider, K.; Kudoh, T.; and Ton,
  V.
\newblock 1997.
\newblock Matsumoto and ekman's japanese and caucasian facial expressions of
  emotion (jacfee): Reliability data and cross-national differences.
\newblock {\em Journal of Nonverbal behavior} 21(1):3--21.

\bibitem[\protect\citeauthoryear{Blackburn, Bone, and
  Phillips}{2001}]{frvt2000}
Blackburn, D.~M.; Bone, M.; and Phillips, P.~J.
\newblock 2001.
\newblock Face recognition vendor test 2000: evaluation report.
\newblock Technical report, DEFENSE ADVANCED RESEARCH PROJECTS AGENCY ARLINGTON
  VA.

\bibitem[\protect\citeauthoryear{Bledsoe}{1966}]{bledsoe1966model}
Bledsoe, W.~W.
\newblock 1966.
\newblock The model method in facial recognition.
\newblock {\em Panoramic Research Inc., Palo Alto, CA, Rep. PR1} 15(47):2.

\bibitem[\protect\citeauthoryear{Blunt}{2019}]{blunt_s.847_2019}
Blunt, R.
\newblock 2019.
\newblock S.847 - 116th {Congress} (2019-2020): {Commercial} {Facial}
  {Recognition} {Privacy} {Act} of 2019.

\bibitem[\protect\citeauthoryear{Bowers}{2019}]{arizona2478}
Bowers, R.
\newblock 2019.
\newblock Arizona {HB}2478 {\textbar} 2019 {\textbar} {Fifty}-fourth
  {Legislature} 1st {Regular}.

\bibitem[\protect\citeauthoryear{Bridges}{2019}]{bridges_2019}
Bridges, E.
\newblock 2019.
\newblock Facial recognition tech is creeping into our lives – i'm going to
  court to stop it.

\bibitem[\protect\citeauthoryear{Buolamwini and Gebru}{2018a}]{Joy1}
Buolamwini, J., and Gebru, T.
\newblock 2018a.
\newblock Gender shades: Intersectional accuracy disparities in commercial
  gender classification.
\newblock In {\em Proc. of the Conference on Fairness, Accountability, and
  Transparency (FAT)}.

\bibitem[\protect\citeauthoryear{Buolamwini and
  Gebru}{2018b}]{buolamwini2018gender}
Buolamwini, J., and Gebru, T.
\newblock 2018b.
\newblock Gender shades: Intersectional accuracy disparities in commercial
  gender classification.
\newblock In {\em Conference on fairness, accountability and transparency},
  77--91.
\newblock PMLR.

\bibitem[\protect\citeauthoryear{Cao \bgroup et al\mbox.\egroup
  }{2018a}]{VGGface2}
Cao, Q.; Shen, L.; Xie, W.; Parkhi, O.~M.; and Zisserman, A.
\newblock 2018a.
\newblock Vggface2: A dataset for recognising faces across pose and age.
\newblock In {\em International Conference on Automatic Face and Gesture
  Recognition}.

\bibitem[\protect\citeauthoryear{Cao \bgroup et al\mbox.\egroup
  }{2018b}]{cao2018vggface2}
Cao, Q.; Shen, L.; Xie, W.; Parkhi, O.~M.; and Zisserman, A.
\newblock 2018b.
\newblock Vggface2: A dataset for recognising faces across pose and age.
\newblock In {\em 2018 13th IEEE International Conference on Automatic Face \&
  Gesture Recognition (FG 2018)},  67--74.
\newblock IEEE.

\bibitem[\protect\citeauthoryear{Cao, Li, and Zhang}{2018}]{cao2018celeb}
Cao, J.; Li, Y.; and Zhang, Z.
\newblock 2018.
\newblock Celeb-500k: A large training dataset for face recognition.
\newblock In {\em 2018 25th IEEE International Conference on Image Processing
  (ICIP)},  2406--2410.
\newblock IEEE.

\bibitem[\protect\citeauthoryear{Carlyle \bgroup et al\mbox.\egroup
  }{2019}]{washingon_5376}
Carlyle; Palumbo; Wellman; Mullet; Pedersen; Billig; Hunt; Liias; Rolfes;
  Saldaña; Hasegawa; and Keiser.
\newblock 2019.
\newblock Washington {State} {Legislature}.

\bibitem[\protect\citeauthoryear{Castro \bgroup et al\mbox.\egroup
  }{2019}]{illinois_sb1719}
Castro, C.; Holmes, L.; Bush, M.; Collins, J.~Y.; Peters, R.; and Murphy, L.~M.
\newblock 2019.
\newblock Illinois {General} {Assembly} - {Bill} {Status} for {SB}1719.

\bibitem[\protect\citeauthoryear{Cavenaugh}{2019}]{arkansas-1943}
Cavenaugh, F.
\newblock 2019.
\newblock Arkansas {HB}1943 {\textbar} 2019 {\textbar} 92nd {General}
  {Assembly}.

\bibitem[\protect\citeauthoryear{Chau}{2019}]{ab1281-privacy}
Chau.
\newblock 2019.
\newblock Bill {Text} - {AB}-1281 {Privacy}: facial recognition technology:
  disclosure.

\bibitem[\protect\citeauthoryear{Chen \bgroup et al\mbox.\egroup
  }{2017}]{chen2017spoofing}
Chen, C.; Dantcheva, A.; Swearingen, T.; and Ross, A.
\newblock 2017.
\newblock Spoofing faces using makeup: An investigative study.
\newblock In {\em 2017 IEEE International Conference on Identity, Security and
  Behavior Analysis (ISBA)},  1--8.
\newblock IEEE.

\bibitem[\protect\citeauthoryear{Chen, Chen, and Hsu}{2015}]{age-bias-mitigate}
Chen, B.-C.; Chen, C.-S.; and Hsu, W.~H.
\newblock 2015.
\newblock Face recognition and retrieval using cross-age reference coding with
  cross-age celebrity dataset.
\newblock {\em IEEE Transactions on Multimedia} 17:804--815.

\bibitem[\protect\citeauthoryear{Chinoy}{2019}]{chinoy_2019}
Chinoy, S.
\newblock 2019.
\newblock We built an 'unbelievable' (but legal) facial recognition machine.

\bibitem[\protect\citeauthoryear{Clarke}{2019}]{clarke_text_2019}
Clarke, Y.~D.
\newblock 2019.
\newblock Text - {H}.{R}.4008 - 116th {Congress} (2019-2020): {No} {Biometric}
  {Barriers} to {Housing} {Act} of 2019.

\bibitem[\protect\citeauthoryear{Coolfire}{2019}]{coolfire_2019}
Coolfire.
\newblock 2019.
\newblock How to improve your crowd control strategy with smart crowd
  monitoring.

\bibitem[\protect\citeauthoryear{{Council of the City of
  Berkeley}}{2018}]{noauthor_surveillance_2018}
{Council of the City of Berkeley}.
\newblock 2018.
\newblock Surveillance {Technology} {Use} and {Community} {Safety} {Ordinance}.

\bibitem[\protect\citeauthoryear{Creem \bgroup et al\mbox.\egroup
  }{2019}]{masachusetts_1385}
Creem, C.~S.; Lewis, J.~P.; Robinson, M.~D.; and Stanley, T.~M.
\newblock 2019.
\newblock Bill {S}.1385.

\bibitem[\protect\citeauthoryear{Dantcheva, Chen, and
  Ross}{2012}]{dantcheva2012can}
Dantcheva, A.; Chen, C.; and Ross, A.
\newblock 2012.
\newblock Can facial cosmetics affect the matching accuracy of face recognition
  systems?
\newblock In {\em 2012 IEEE Fifth international conference on biometrics:
  theory, applications and systems (BTAS)},  391--398.
\newblock IEEE.

\bibitem[\protect\citeauthoryear{Dearden}{2019}]{dearden_2019}
Dearden, L.
\newblock 2019.
\newblock Facial recognition wrongly identifies public criminals 96\% of time,
  figures reveal.

\bibitem[\protect\citeauthoryear{Durkin}{2019}]{durkin_2019}
Durkin, E.
\newblock 2019.
\newblock New york tenants fight as landlords embrace facial recognition
  cameras.

\bibitem[\protect\citeauthoryear{Eisenthal, Dror, and
  Ruppin}{2006}]{eisenthal_facial_2006}
Eisenthal, Y.; Dror, G.; and Ruppin, E.
\newblock 2006.
\newblock Facial attractiveness: beauty and the machine.
\newblock {\em Neural Computation} 18(1):119--142.

\bibitem[\protect\citeauthoryear{Farmer}{2019}]{florida-1270}
Farmer, G.
\newblock 2019.
\newblock {FL} - {S}1270.

\bibitem[\protect\citeauthoryear{Fetscherin, Tantleff-Dunn, and
  Klumb}{2019}]{fetscherin_effects_2019}
Fetscherin, M.; Tantleff-Dunn, S.; and Klumb, A.
\newblock 2019.
\newblock Effects of facial features and styling elements on perceptions of
  competence, warmth, and hireability of male professionals.
\newblock {\em The Journal of Social Psychology} 0(0):1--14.

\bibitem[\protect\citeauthoryear{Flanagan}{2015}]{flanagan2015chexia}
Flanagan, P.~A.
\newblock 2015.
\newblock Chexia face recognition.

\bibitem[\protect\citeauthoryear{Forczma{\'n}ski and
  Furman}{2012}]{forczmanski2012comparative}
Forczma{\'n}ski, P., and Furman, M.
\newblock 2012.
\newblock Comparative analysis of benchmark datasets for face recognition
  algorithms verification.
\newblock In {\em International Conference on Computer Vision and Graphics},
  354--362.
\newblock Springer.

\bibitem[\protect\citeauthoryear{Founds \bgroup et al\mbox.\egroup
  }{2011}]{founds2011nist}
Founds, A.~P.; Orlans, N.; Genevieve, W.; and Watson, C.~I.
\newblock 2011.
\newblock Nist special databse 32-multiple encounter dataset ii (meds-ii).
\newblock Technical report.

\bibitem[\protect\citeauthoryear{Gebru \bgroup et al\mbox.\egroup
  }{2018}]{gebru2018datasheets}
Gebru, T.; Morgenstern, J.; Vecchione, B.; Vaughan, J.~W.; Wallach, H.;
  Daume{\'e}~III, H.; and Crawford, K.
\newblock 2018.
\newblock Datasheets for datasets.
\newblock {\em arXiv preprint arXiv:1803.09010}.

\bibitem[\protect\citeauthoryear{Grgic, Delac, and
  Grgic}{2011}]{grgic2011scface}
Grgic, M.; Delac, K.; and Grgic, S.
\newblock 2011.
\newblock Scface--surveillance cameras face database.
\newblock {\em Multimedia tools and applications} 51(3):863--879.

\bibitem[\protect\citeauthoryear{Grother, Ngan, and
  Hanaoka}{2018}]{frvtongoing}
Grother, P.~J.; Ngan, M.~L.; and Hanaoka, K.~K.
\newblock 2018.
\newblock Ongoing face recognition vendor test (frvt) part 2: Identification.
\newblock Technical report.

\bibitem[\protect\citeauthoryear{Guo \bgroup et al\mbox.\egroup
  }{2016}]{guo2016ms}
Guo, Y.; Zhang, L.; Hu, Y.; He, X.; and Gao, J.
\newblock 2016.
\newblock Ms-celeb-1m: A dataset and benchmark for large-scale face
  recognition.
\newblock In {\em European Conference on Computer Vision},  87--102.
\newblock Springer.

\bibitem[\protect\citeauthoryear{Han \bgroup et al\mbox.\egroup
  }{2017}]{han2017heterogeneous}
Han, H.; Jain, A.~K.; Wang, F.; Shan, S.; and Chen, X.
\newblock 2017.
\newblock Heterogeneous face attribute estimation: A deep multi-task learning
  approach.
\newblock {\em IEEE transactions on pattern analysis and machine intelligence}
  40(11):2597--2609.

\bibitem[\protect\citeauthoryear{Harvey and LaPlace}{2020}]{megapixels}
Harvey, A., and LaPlace, J.
\newblock 2020.
\newblock Megapixels: Origins and endpoints of datasets created "in the wild".

\bibitem[\protect\citeauthoryear{Hasegawa, Saldaña, and
  Nguyen}{2019}]{washington_5528}
Hasegawa; Saldaña; and Nguyen.
\newblock 2019.
\newblock Washington {State} {Legislature}.

\bibitem[\protect\citeauthoryear{Haskins}{2019}]{haskins_oakland_2019}
Haskins, C.
\newblock 2019.
\newblock Oakland {Becomes} {Third} {U}.{S}. {City} to {Ban} {Facial}
  {Recognition}.

\bibitem[\protect\citeauthoryear{Huang \bgroup et al\mbox.\egroup
  }{2007}]{LFWTech}
Huang, G.~B.; Ramesh, M.; Berg, T.; and Learned-Miller, E.
\newblock 2007.
\newblock Labeled faces in the wild: A database for studying face recognition
  in unconstrained environments.
\newblock Technical Report 07-49, University of Massachusetts, Amherst.

\bibitem[\protect\citeauthoryear{{Idaho Judiciary, Rules and Administration
  Committee}}{2019}]{idaho-0118}
{Idaho Judiciary, Rules and Administration Committee}.
\newblock 2019.
\newblock Idaho {H}0118 {\textbar} 2019 {\textbar} {Regular} {Session}.

\bibitem[\protect\citeauthoryear{{Illinois Legislature}}{2008}]{illinois-740}
{Illinois Legislature}.
\newblock 2008.
\newblock 740 {ILCS} 14/  {Biometric} {Information} {Privacy} {Act}.

\bibitem[\protect\citeauthoryear{Irani and
  Silberman}{2013}]{irani2013turkopticon}
Irani, L.~C., and Silberman, M.~S.
\newblock 2013.
\newblock Turkopticon: Interrupting worker invisibility in amazon mechanical
  turk.
\newblock In {\em Proceedings of the SIGCHI conference on human factors in
  computing systems},  611--620.

\bibitem[\protect\citeauthoryear{Jafri and Arabnia}{2009}]{jafri2009survey}
Jafri, R., and Arabnia, H.~R.
\newblock 2009.
\newblock A survey of face recognition techniques.
\newblock {\em Jips} 5(2):41--68.

\bibitem[\protect\citeauthoryear{Jain and Learned-Miller}{2010}]{jain2010fddb}
Jain, V., and Learned-Miller, E.
\newblock 2010.
\newblock Fddb: A benchmark for face detection in unconstrained settings.

\bibitem[\protect\citeauthoryear{Jain and Li}{2011}]{jain2011handbook}
Jain, A.~K., and Li, S.~Z.
\newblock 2011.
\newblock {\em Handbook of face recognition}.
\newblock Springer.

\bibitem[\protect\citeauthoryear{Kelley \bgroup et al\mbox.\egroup
  }{2009}]{kelley2009nutrition}
Kelley, P.~G.; Bresee, J.; Cranor, L.~F.; and Reeder, R.~W.
\newblock 2009.
\newblock A nutrition label for privacy.
\newblock In {\em Proceedings of the 5th Symposium on Usable Privacy and
  Security}, ~4.
\newblock ACM.

\bibitem[\protect\citeauthoryear{Kemelmacher-Shlizerman \bgroup et
  al\mbox.\egroup }{2016}]{kemelmacher2016megaface}
Kemelmacher-Shlizerman, I.; Seitz, S.~M.; Miller, D.; and Brossard, E.
\newblock 2016.
\newblock The megaface benchmark: 1 million faces for recognition at scale.
\newblock In {\em Proceedings of the IEEE Conference on Computer Vision and
  Pattern Recognition},  4873--4882.

\bibitem[\protect\citeauthoryear{Keyes}{2018a}]{keyes_misgendering_2018}
Keyes, O.
\newblock 2018a.
\newblock The {Misgendering} {Machines}: {Trans}/{HCI} {Implications} of
  {Automatic} {Gender} {Recognition}.
\newblock {\em Proceedings of the ACM on Human-Computer Interaction}
  2(CSCW):1--22.

\bibitem[\protect\citeauthoryear{Keyes}{2018b}]{Keyes:2018:MMT:3290265.3274357}
Keyes, O.
\newblock 2018b.
\newblock The misgendering machines: Trans/{HCI} implications of automatic
  gender recognition.
\newblock {\em Proc. of the Human Computer Interact action} 2(CSCW).

\bibitem[\protect\citeauthoryear{Lazarus, Gupta, and
  Panda}{2018}]{indian_face_database2018}
Lazarus, M.~Z.; Gupta, S.; and Panda, N.
\newblock 2018.
\newblock An indian facial database highlighting the spectacle problems.

\bibitem[\protect\citeauthoryear{Leuner}{2019}]{leuner_replication_2019}
Leuner, J.
\newblock 2019.
\newblock A {Replication} {Study}: {Machine} {Learning} {Models} {Are}
  {Capable} of {Predicting} {Sexual} {Orientation} {From} {Facial} {Images}.
\newblock {\em arXiv:1902.10739 [cs]}.
\newblock arXiv: 1902.10739.

\bibitem[\protect\citeauthoryear{Li \bgroup et al\mbox.\egroup
  }{2017}]{asian_celeb_dataset}
Li, D.; Zhang, X.; Song, L.; and Zhao, Y.
\newblock 2017.
\newblock Multiple-step model training for face recognition.
\newblock In {\em International Conference on Applications and Techniques in
  Cyber Security and Intelligence},  146--153.
\newblock Springer.

\bibitem[\protect\citeauthoryear{Liu \bgroup et al\mbox.\egroup
  }{2015}]{CelebA}
Liu, Z.; Luo, P.; Wang, X.; and Tang, X.
\newblock 2015.
\newblock Deep learning face attributes in the wild.
\newblock In {\em Proceedings of International Conference on Computer Vision
  (ICCV)}.

\bibitem[\protect\citeauthoryear{LoBue and Thrasher}{2015}]{lobue2015child}
LoBue, V., and Thrasher, C.
\newblock 2015.
\newblock The child affective facial expression (cafe) set: Validity and
  reliability from untrained adults.
\newblock {\em Frontiers in psychology} 5:1532.

\bibitem[\protect\citeauthoryear{Lucido}{2019}]{michigan0342}
Lucido, P.~J.
\newblock 2019.
\newblock Michigan {SB}0342 {\textbar} 2019-2020 {\textbar} 100th
  {Legislature}.

\bibitem[\protect\citeauthoryear{Manthorpe and
  Martin}{2019}]{manthorpe_martin_2019}
Manthorpe, R., and Martin, A.~J.
\newblock 2019.
\newblock 81\% of 'suspects' flagged by met's police facial recognition
  technology innocent, independent report says.

\bibitem[\protect\citeauthoryear{Menegus}{2019}]{gizmodo}
Menegus, B.
\newblock 2019.
\newblock Defense of amazon's face recognition tool undermined by its only
  known police client.

\bibitem[\protect\citeauthoryear{{Merler} \bgroup et al\mbox.\egroup
  }{2019}]{2019arXiv190110436M}
{Merler}, M.; {Ratha}, N.; {Feris}, R.~S.; and {Smith}, J.~R.
\newblock 2019.
\newblock {Diversity in Faces}.
\newblock {\em arXiv preprints}  arXiv:1901.10436.

\bibitem[\protect\citeauthoryear{Mesnik}{2018}]{mesnik_2018}
Mesnik, B.
\newblock 2018.
\newblock How face recognition works in a crowd.

\bibitem[\protect\citeauthoryear{Metz}{2019}]{metz_facial_2019}
Metz, C.
\newblock 2019.
\newblock Facial {Recognition} {Tech} {Is} {Growing} {Stronger}, {Thanks} to
  {Your} {Face}.
\newblock {\em The New York Times}.

\bibitem[\protect\citeauthoryear{Mitchell \bgroup et al\mbox.\egroup
  }{2018}]{ModCards}
Mitchell, M.; Wu, S.; Zaldivar, A.; Barnes, P.; Vasserman, L.; Hutchinson, B.;
  Spitzer, E.; Raji, I.~D.; and Gebru, T.
\newblock 2018.
\newblock Model cards for model reporting.
\newblock {\em CoRR} abs/1810.03993.

\bibitem[\protect\citeauthoryear{Montgomery and
  Hagemann}{2019}]{montgomery_precision_nodate}
Montgomery, C., and Hagemann, R.
\newblock 2019.
\newblock Precision {Regulation} and {Facial} {Recognition}.
\newblock ~4.

\bibitem[\protect\citeauthoryear{Morales, Fierrez, and
  Vera-Rodriguez}{2019}]{morales2019sensitivenets}
Morales, A.; Fierrez, J.; and Vera-Rodriguez, R.
\newblock 2019.
\newblock Sensitivenets: Learning agnostic representations with application to
  face recognition.
\newblock {\em arXiv preprint arXiv:1902.00334} 1(2).

\bibitem[\protect\citeauthoryear{Ngan and Grother}{}]{ngan_grother}
Ngan, M., and Grother, P.
\newblock Face recognition vendor test (frvt) performance of automated gender
  classification algorithms.

\bibitem[\protect\citeauthoryear{Ngan, Ngan, and Grother}{2015}]{FRVT2015}
Ngan, M.; Ngan, M.; and Grother, P.
\newblock 2015.
\newblock Face recognition vendor test {(FRVT)} performance of automated gender
  classification algorithms.
\newblock Government technical report, US Department of Commerce, National
  Institute of Standards and Technology.

\bibitem[\protect\citeauthoryear{NIST}{2015}]{chexia_2015}
NIST.
\newblock 2015.
\newblock Chexia {Face} {Recognition}.

\bibitem[\protect\citeauthoryear{NIST}{2018}]{nist_2018}
NIST.
\newblock 2018.
\newblock Nist evaluation shows advance in face recognition software's
  capabilities.

\bibitem[\protect\citeauthoryear{O'Flaherty}{2019}]{oflaherty_2019}
O'Flaherty, K.
\newblock 2019.
\newblock Facial recognition at u.s. airports. should you be concerned?

\bibitem[\protect\citeauthoryear{Parkhi \bgroup et al\mbox.\egroup
  }{2015}]{vgg}
Parkhi, O.~M.; Vedaldi, A.; Zisserman, A.; et~al.
\newblock 2015.
\newblock Deep face recognition.
\newblock In {\em bmvc}, volume~1, ~6.

\bibitem[\protect\citeauthoryear{Phillips \bgroup et al\mbox.\egroup
  }{2000a}]{phillips2000introduction}
Phillips, P.~J.; Martin, A.; Wilson, C.~L.; and Przybocki, M.
\newblock 2000a.
\newblock An introduction evaluating biometric systems.
\newblock {\em Computer} 33(2):56--63.

\bibitem[\protect\citeauthoryear{Phillips \bgroup et al\mbox.\egroup
  }{2000b}]{Ref27}
Phillips, P.; Moon, H.; Rizvi, S.; and Rauss, P.
\newblock 2000b.
\newblock The feret evaluation methodology for face-recognition algorithms.
\newblock In {\em IEEE Transactions on Pattern Analysis and Machine
  Intelligence}, volume~22,  1090 -- 1104.
\newblock IEEE.

\bibitem[\protect\citeauthoryear{Phillips \bgroup et al\mbox.\egroup
  }{2003}]{frvt2002}
Phillips, P.~J.; Grother, P.~J.; Micheals, R.~J.; Blackburn, D.~M.; Tabassi,
  E.; and Bone, M.
\newblock 2003.
\newblock Face recognition vendor test 2002: Evaluation report.
\newblock Technical report.

\bibitem[\protect\citeauthoryear{Phillips \bgroup et al\mbox.\egroup
  }{2005}]{FRGC}
Phillips, P.~J.; Flynn, P.~J.; Scruggs, T.; Bowyer, K.~W.; Chang, J.; Hoffman,
  K.; Marques, J.; Min, J.; and Worek, W.
\newblock 2005.
\newblock Overview of the face recognition grand challenge.
\newblock In {\em 2005 IEEE computer society conference on computer vision and
  pattern recognition (CVPR'05)}, volume~1,  947--954.
\newblock IEEE.

\bibitem[\protect\citeauthoryear{Picard}{2000}]{picard_affective_2000}
Picard, R.~W.
\newblock 2000.
\newblock {\em Affective {Computing}}.
\newblock MIT Press.
\newblock Google-Books-ID: GaVncRTcb1gC.

\bibitem[\protect\citeauthoryear{Raji and Buolamwini}{2019}]{AA}
Raji, I.~D., and Buolamwini, J.
\newblock 2019.
\newblock Actionable auditing: Investigating the impact of publicly naming
  biased performance results of commercial {AI} products.
\newblock In {\em Prof. of the Conference on Artificial Intelligence, Ethics,
  and Society}.

\bibitem[\protect\citeauthoryear{Ristani \bgroup et al\mbox.\egroup
  }{2016}]{duke_ristani2016performance}
Ristani, E.; Solera, F.; Zou, R.; Cucchiara, R.; and Tomasi, C.
\newblock 2016.
\newblock Performance measures and a data set for multi-target, multi-camera
  tracking.
\newblock In {\em European Conference on Computer Vision},  17--35.
\newblock Springer.

\bibitem[\protect\citeauthoryear{Ritchie}{2019}]{newyork-assembly}
Ritchie.
\newblock 2019.
\newblock New {York} {State} {Assembly} {\textbar} {Bill} {Search} and
  {Legislative} {Information}.

\bibitem[\protect\citeauthoryear{Schmid, Marx, and
  Samal}{2008}]{schmid_computation_2008}
Schmid, K.; Marx, D.; and Samal, A.
\newblock 2008.
\newblock Computation of a face attractiveness index based on neoclassical
  canons, symmetry, and golden ratios.
\newblock {\em Pattern Recognition} 41(8):2710--2717.

\bibitem[\protect\citeauthoryear{Sharif \bgroup et al\mbox.\egroup
  }{2017}]{sharif2017face}
Sharif, M.; Naz, F.; Yasmin, M.; Shahid, M.~A.; and Rehman, A.
\newblock 2017.
\newblock Face recognition: A survey.
\newblock {\em Journal of Engineering Science \& Technology Review} 10(2).

\bibitem[\protect\citeauthoryear{Shultz}{2019}]{shultz_2019}
Shultz, J.
\newblock 2019.
\newblock Spying on children won't keep them safe.

\bibitem[\protect\citeauthoryear{Snow}{2018}]{ref48}
Snow, J.
\newblock 2018.
\newblock Amazon’s face recognition falsely matched 28 members of congress
  with mugshots.

\bibitem[\protect\citeauthoryear{{Somerville City
  Council}}{2019}]{noauthor_ordinance_nodate}
{Somerville City Council}.
\newblock 2019.
\newblock Ordinance {No}. 2019-16 {\textbar} {Code} of {Ordinances} {\textbar}
  {Somerville}, {MA} {\textbar} {Municode} {Library}.

\bibitem[\protect\citeauthoryear{Spears}{2019}]{spears_2019}
Spears, M.
\newblock 2019.
\newblock 'look at camera for entry': Tacoma convenience store using facial
  recognition technology.

\bibitem[\protect\citeauthoryear{Stewart, Andriluka, and
  Ng}{2016}]{brainwash_stewart2016end}
Stewart, R.; Andriluka, M.; and Ng, A.~Y.
\newblock 2016.
\newblock End-to-end people detection in crowded scenes.
\newblock In {\em Proceedings of the IEEE conference on computer vision and
  pattern recognition},  2325--2333.

\bibitem[\protect\citeauthoryear{Taigman \bgroup et al\mbox.\egroup
  }{2014}]{taigman2014deepface}
Taigman, Y.; Yang, M.; Ranzato, M.; and Wolf, L.
\newblock 2014.
\newblock Deepface: Closing the gap to human-level performance in face
  verification.
\newblock In {\em Proceedings of the IEEE conference on computer vision and
  pattern recognition},  1701--1708.

\bibitem[\protect\citeauthoryear{{Texas
  Legislature}}{2019}]{texas_business_commerce}
{Texas Legislature}.
\newblock 2019.
\newblock {Business} {and} {Commerce} {Code} {Chapter} 503. {Biometric}
  {Identifiers}.

\bibitem[\protect\citeauthoryear{Ting}{2019}]{ab1215-facial}
Ting.
\newblock 2019.
\newblock Bill {Text} - {AB}-1215 {Law} enforcement: facial recognition and
  other biometric surveillance.

\bibitem[\protect\citeauthoryear{Vangara \bgroup et al\mbox.\egroup
  }{2019}]{vangara2019characterizing}
Vangara, K.; King, M.~C.; Albiero, V.; Bowyer, K.; et~al.
\newblock 2019.
\newblock Characterizing the variability in face recognition accuracy relative
  to race.
\newblock In {\em Proceedings of the IEEE Conference on Computer Vision and
  Pattern Recognition Workshops},  0--0.

\bibitem[\protect\citeauthoryear{Wallace \bgroup et al\mbox.\egroup
  }{2019}]{nys_biometrics_bill}
Wallace; Epstein; Mosley; MG, M.; Simon; Gottfried; L, R.; Reyes; Otis;
  Simotas; Quart; Kim; Rodriguez; Fahy; Abinanti; and Weprin.
\newblock 2019.
\newblock New {York} {State} {Assembly} {\textbar} {Bill} {Search} and
  {Legislative} {Information}.

\bibitem[\protect\citeauthoryear{Wang and Deng}{2018}]{wang2018deep}
Wang, M., and Deng, W.
\newblock 2018.
\newblock Deep face recognition: A survey.
\newblock {\em arXiv preprint arXiv:1804.06655}.

\bibitem[\protect\citeauthoryear{Wang and Kosinski}{}]{wang_kosinski}
Wang, Yilun~Wang, M.~K., and Kosinski, M.
\newblock Deep neural networks can detect sexual orientation from face.

\bibitem[\protect\citeauthoryear{Wang \bgroup et al\mbox.\egroup
  }{2018}]{Wang2018RacialFI}
Wang, M.; Deng, W.; Hu, J.; Peng, J.; Tao, X.; and Huang, Y.
\newblock 2018.
\newblock Racial faces in-the-wild: Reducing racial bias by deep unsupervised
  domain adaptation.
\newblock {\em CoRR} abs/1812.00194.

\bibitem[\protect\citeauthoryear{Wong \bgroup et al\mbox.\egroup
  }{2011}]{wong2011patch}
Wong, Y.; Chen, S.; Mau, S.; Sanderson, C.; and Lovell, B.~C.
\newblock 2011.
\newblock Patch-based probabilistic image quality assessment for face selection
  and improved video-based face recognition.
\newblock In {\em CVPR 2011 WORKSHOPS},  74--81.
\newblock IEEE.

\bibitem[\protect\citeauthoryear{Wu and Zhang}{2016}]{wu_automated_2016}
Wu, X., and Zhang, X.
\newblock 2016.
\newblock Automated {Inference} on {Criminality} using {Face} {Images}.
\newblock {\em ArXiv} abs/1611.04135.
\newblock arXiv: 1611.04135.

\bibitem[\protect\citeauthoryear{Yang, Kriegman, and
  Ahuja}{2002}]{yang2002detecting}
Yang, M.-H.; Kriegman, D.~J.; and Ahuja, N.
\newblock 2002.
\newblock Detecting faces in images: A survey.
\newblock {\em IEEE Transactions on pattern analysis and machine intelligence}
  24(1):34--58.

\bibitem[\protect\citeauthoryear{Yee and Ronen}{2019}]{administrative_code}
Yee, W., and Ronen, H.
\newblock 2019.
\newblock Acquisition of surveillance technology.

\bibitem[\protect\citeauthoryear{Yi \bgroup et al\mbox.\egroup
  }{2014}]{yi2014learningcasia}
Yi, D.; Lei, Z.; Liao, S.; and Li, S.~Z.
\newblock 2014.
\newblock Learning face representation from scratch.
\newblock {\em arXiv preprint arXiv:1411.7923}.

\bibitem[\protect\citeauthoryear{Zhao \bgroup et al\mbox.\egroup
  }{2003}]{zhao2003face}
Zhao, W.; Chellappa, R.; Phillips, P.~J.; and Rosenfeld, A.
\newblock 2003.
\newblock Face recognition: A literature survey.
\newblock {\em ACM computing surveys (CSUR)} 35(4):399--458.

\end{thebibliography}
\end{small}
\end{quote}

\end{document}